\documentclass{article}

\usepackage{spconf}
\usepackage{amsmath}
\usepackage{amssymb}
\usepackage{graphicx}
\usepackage{hyperref}
\usepackage{multirow}
\usepackage{booktabs}
\usepackage{xcolor}

\title{SmoGVLM: A Small, Graph-Enhanced Vision-Language Model}

\name{Debjyoti Mondal, Rituraj Singh, Subhadarshi Panda}
\address{Samsung R\&D Institute India-Bangalore}

\begin{document}

\maketitle

\begin{abstract}
Large vision-language models (VLMs) achieve strong performance on multimodal tasks but often suffer from hallucination and poor grounding in knowledge-intensive reasoning. 
We propose \textbf{SmoGVLM}, a \emph{small, graph-enhanced VLM} that integrates structured knowledge with visual and textual modalities, using Graph Neural Networks.
We investigate the effects of our method across a range of model sizes, from tiny ($1.3$B) to large ($13$B) models.
Our results demonstrate that, when trained using our approach, a small model can achieve performance gains upto $16.24\%$, and surpass its larger counterparts, outperforming larger VLMs and strong fine-tuned baselines.
These results highlight the potential of structured knowledge augmentation for efficient, smaller-scale multimodal reasoning systems.
\end{abstract}


\section{Introduction}
\label{sec:intro}
Large vision-language models (VLMs) have achieved impressive performance across a wide range of multimodal tasks, from visual question answering (VQA) to reasoning over images and text \cite{liu2023visual, li2023blip}.
However, these models often suffer from hallucinations and poor grounding when faced with knowledge-intensive queries. 
It is especially problematic for smaller models, which lack sufficient capacity to internalize factual world knowledge.
Scaling model size offers improvements, but incurs prohibitive costs in training, inference, and deployment.
Knowledge graphs (KGs) provide structured context that can guide reasoning and reduce hallucination.

In this work, we propose SmoGVLM, a small graph-enhanced VLM that integrates structured knowledge with vision and language representations.
We extract compact sub-graphs from ConceptNet \cite{speer2017conceptnet} and encode them using a lightweight Graph Neural Network (GNN), which are then fused with image and text embeddings.
This enables the model to reason in a chain-of-thought style while remaining efficient.
Crucially, our sub-graph extraction is simple and fast, avoiding the excessive cost of prior methods such as QA-GNN \cite{yasunaga2021qa}, which rank every concept against the query.

Our contributions are threefold:
\begin{enumerate}
    \item An efficient sub-graph extraction method that avoids expensive ranking, yet improves speed and relevance.
    \item A lightweight projection-based technique that facilitates the fusion of image, language and KGs.
    \item SmoGVLM, a small, graph-enhanced VLM for knowledge intensive VQA.
\end{enumerate}

We evaluate SmoGVLM's performance on ScienceQA \cite{lu2022learn} and A-OKVQA \cite{aokvqa}, both requiring multimodal reasoning and external knowledge.
Results show that even a $1.3$B SmoGVLM significantly outperforms larger VLMs such as LLaVA-$7$B by $5.8\%$.
These findings establish that structured KG augmentation enables smaller VLMs to rival larger models while reducing hallucinations and compute costs.

\section{Related Work}
\label{sec:related_work}

\subsection{Vision-Language Models}
Recent VLMs align large language models with visual encoders for multimodal reasoning.
Examples include BLIP-2 \cite{li2023blip}, LLaVA \cite{liu2023visual}, InstructBLIP \cite{dai2023instructblip}, and MiniGPT-4 \cite{zhu2023minigpt}.
They achieve strong results on tasks like captioning and VQA, but often hallucinate on knowledge-intensive queries \cite{liu2024survey}.
Scaling model size reduces some of these issues, but then training and inference costs increase.

\subsection{Knowledge Augmented Reasoning}
Prior work has explored integrating external knowledge into language and multimodal models.
QA-GNN \cite{yasunaga2021qa} and KAM-CoT \cite{mondal2024kam} leverage knowledge graphs for QA, but their extraction strategies rank a large number of concepts per query, leading to high complexity.
Other approaches verbalize KG triples \cite{wang2024knowledge} and inject them as text, but they miss the graph structure entirely.
In contrast, our method directly encodes compact sub-graphs, combining efficiency with structure.

\subsection{GNNs for Knowledge Integration}
GNNs are widely used for encoding structured knowledge \cite{schlichtkrull2018modeling}
and have also improved reasoning in knowledge augmented NLP tasks \cite{lin2019kagnet, yasunaga2022deep}.
In multimodal settings, however, they are rarely combined with large VLMs due to scalability concerns.
We show that lightweight sub-graph encoding can be integrated into VLMs, yielding strong gains without heavy compute overhead.

\section{Method}
\label{sec:method}

\subsection{Task formulation}
We consider the task of multimodal question answering, where a question $q$ with answer options $\{a_1, a_2, \ldots, a_k\}$ is given along with an image $X_{\text{img}}$ and optional textual context $c$. 
The objective is to generate both a rationale $r$ and a final answer $\hat{a}$. 
Formally, given inputs $X_{\text{lang}}, X_{\text{img}}, X_{\text{kg}}$, the model learns to maximize the likelihood of generating reference text $Y=[r;\hat{a}]$ of length $N$:
\begin{equation*}
p(Y|X_{\text{lang}},X_{\text{img}},X_{\text{kg}}) = \prod_{i=1}^{N} p_\theta(Y_i|X_{\text{lang}},X_{\text{img}},X_{\text{kg}},Y_{<i}),
\end{equation*}
where $\theta$ denotes model parameters.
This encourages the model to first generate reasoning steps and then the answer.

\subsection{Encoding image and text modalities}
We begin by encoding language and visual inputs into a shared $d$-dimensional space. 
For text input $X_{\text{lang}}$, we use a trainable embedding layer:
\begin{align}
H_{\text{lang}} &= \text{Tokenizer\&Embedder}(X_{\text{lang}}) \in \mathbb{R}^{n \times d},
\end{align}
where $n$ is the number of tokens. 
For image input $X_{\text{img}}$, we use an encoder followed by a trainable linear projection:
\begin{align}
H_{\text{img}} &= \text{ImageEncoder}(X_{\text{img}}) W_{\text{img}} \in \mathbb{R}^{m \times d},
\end{align}
where $m$ is the number of image patches. 
In our method, we adopt LLaVA \cite{liu2023visual} as the backbone, with LLaMA \cite{touvron2023llama} as the language model and CLIP \cite{DBLP:conf/icml/RadfordKHRGASAM21} as our ImageEncoder($\cdot$).

\subsection{Sub-graph extraction}
To incorporate structured knowledge, we extract a relevant sub-graph from ConceptNet \cite{speer2017conceptnet}. 
The complete graph contains $\sim$$800$k English entities and $2$M triples, spanned across $34$ relation types.
We then use BLIP-2 \cite{li2023blip} to compute embeddings of all triples and compare them against multimodal embeddings of the given question-image pair using cosine similarity. 
The top-$k$ ranked triples (up to $200$ nodes) are retained to form a compact sub-graph $X_{\text{kg}}$. 
This simple, lightweight extraction avoids the excessive computational costs of ranking every concept with the question, as required in QA-GNN \cite{yasunaga2021qa} and KAM-CoT \cite{mondal2024kam}. 
We quantify this speedup in Section \ref{sec:experiments}.

\subsection{Encoding the sub-graph}
The sub-graph $X_{\text{kg}}$ is encoded using a stack of GNNs.
For each concept node, we average its span embeddings across all occurrences in the complete KG.
This is a one-time preprocessing step and gives node embeddings with the same dimensionality as the language model.
Relation types are represented using a trainable lookup table of size $34 \times 64$, where $34$ is the number of relation categories.
The KGEncoder($\cdot$) is made of a Relational Graph Attention Network (RGAT) \cite{busbridge2019relational}, followed by a Graph Convolutional Network (GCN) \cite{DBLP:conf/iclr/KipfW17}, with a LeakyReLU activation in between.
\begin{align}
H_{\text{kg}} = \text{KGEncoder}(X_{\text{kg}}) \in \mathbb{R}^{p \times d},
\end{align}
where $p$ is the number of nodes. This representation captures both entity semantics and relational structure.

\subsection{Model overview}
\begin{figure}
    \centering
    \includegraphics[width=\linewidth]{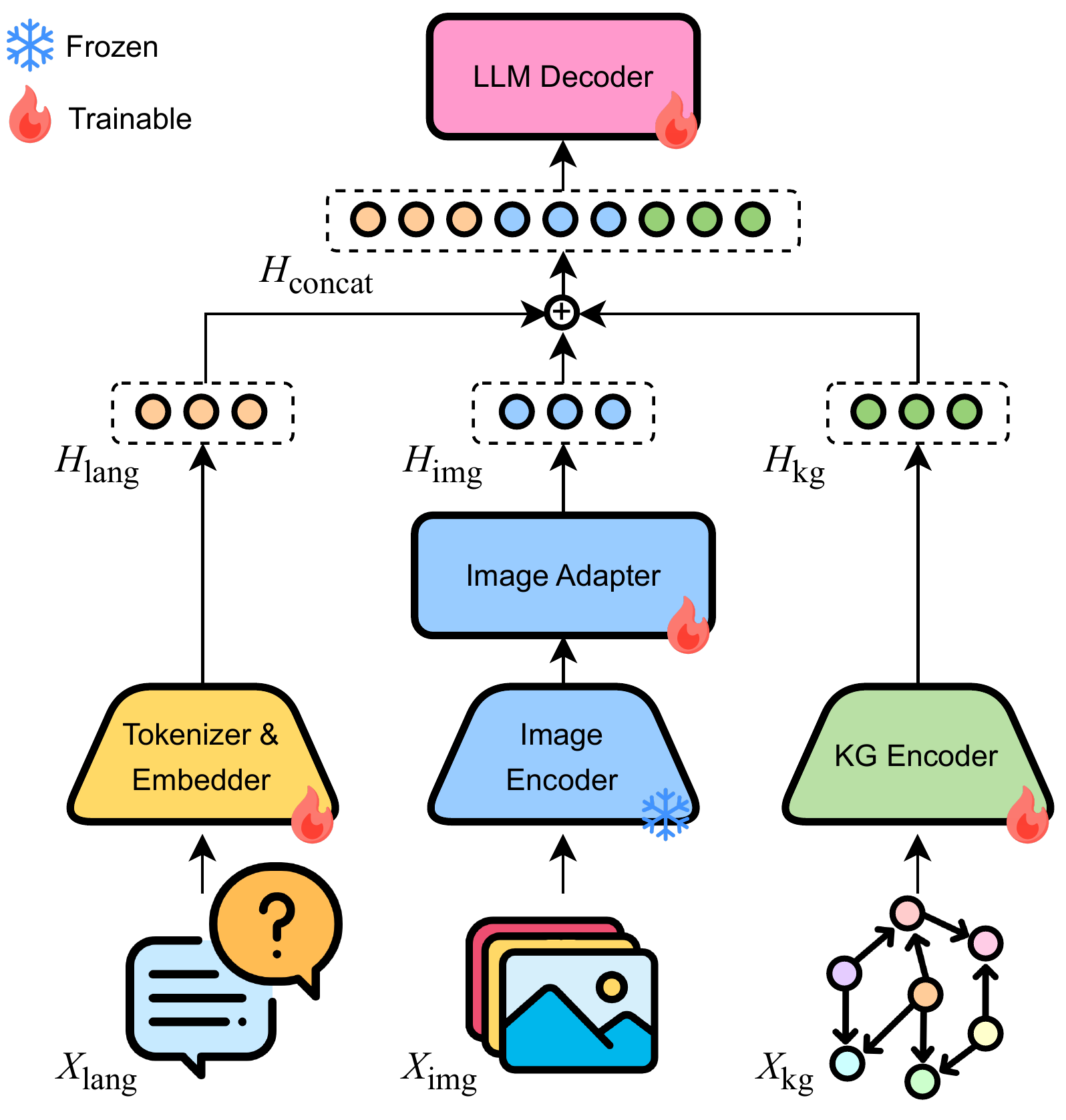}
    \caption{
        Overview of SmoGVLM. 
        Language, image, and KG inputs are encoded, fused into a joint representation, and decoded into rationale and answer.
    }
    \label{fig:architecture}
\end{figure}

Finally, we integrate all modalities. The encoded representations are concatenated:
\begin{align}
H_{\text{concat}} = [H_{\text{kg}}; H_{\text{img}}; H_{\text{lang}}] \in \mathbb{R}^{(p+m+n)\times d}
\end{align}
This joint representation is passed to a transformer decoder, which generates the rationale and answer autoregressively:
\begin{align}
p(Y_t|Y_{<t},X_{\text{lang}},X_{\text{img}},X_{\text{kg}}) = \text{Decoder}(Y_{<t}, H_{\text{concat}})
\end{align}
By conditioning the answer on the generated rationale, the model ensures reasoning consistency and reduces hallucination. 
An overview of the architecture is shown in Fig. \ref{fig:architecture}.

\section{Experiments}
\label{sec:experiments}

\subsection{Datasets}
We evaluate SmoGVLM on two multimodal benchmarks. 

\textbf{ScienceQA} \cite{lu2022learn} contains $21.2$k multiple-choice questions across natural, social, and language sciences, with accompanying text, images, and explanations. 
It comes with $12.7$k train, $4.2$k validation, and $4.2$k test examples. 

\textbf{A-OKVQA} \cite{aokvqa} consists of $25$k open-ended visual questions requiring commonsense and world knowledge, with $17$k training, $1.1$k validation, and $6.7$k test samples. 
Since test labels are not publicly available, we follow prior work and treat the original validation set as test data. 
The training set is further split into $12,726$ train and $4,330$ validation samples.

\subsection{Baselines}
To quantify the effectiveness of adding structured knowledge, we perform experiments in the following settings:
\begin{itemize}
    \item \textbf{No knowledge:} Finetuning a standard VLM with no external knowledge.
    \item \textbf{With structured knowledge:} Incorporating structured sub-graphs into the VLM using GNNs.
\end{itemize}
Experiments are conducted across model sizes from $1.3$B to $13$B parameters, using both full fine-tuning (FFT) and parameter-efficient LoRA. 

For sub-graph extraction, we compare our lightweight top-$k$ retrieval with QA-GNN \cite{yasunaga2021qa}, which ranks every concept against the question, enabling a direct analysis of both accuracy and time required.

\subsection{Training details}
All experiments are performed on a cluster of $8$ A$100$ $40$GB GPUs.
To fit models of sizes $\geq 7$B, we employ FSDP.
We use a learning rate of $2\times10^{-5}$ and cosine decay scheduling. 
For parameter-efficient tuning, we adopt LoRA with $(r, \alpha) = (128, 128)$ and a dropout rate of $0.05$.
All models are trained in \emph{bfloat16} precision. We train for a maximum of $20$ epochs with an early stopping patience of $3$ epochs.

\subsection{Results}
We assess the efficacy of sub-graph extraction methods using two metrics:
(a) \textbf{Time Taken} to extract a relevant sub-graph, and (b) \textbf{Similarity} of extracted triples with the answer.
Table \ref{tab:subgraph} shows that our method is over $10\times$ faster than QA-GNN and closer to the answer by $34\%$.
This efficiency comes from ranking only triples, whereas QA-GNN ranks every grounded concept and its $2$-hop neighbors against the question.

Table \ref{tab:scienceqa} presents results on ScienceQA.
SmoGVLM improves over LLaVA by $12.55\%$, $17.45\%$, and $0.92\%$ for the $1.3$B, $7$B, and $13$B FFT models respectively.
The gains are most pronounced for smaller models ($\leq7$B).
This is expected since larger models are trained on broader corpora.
Notably, SmoGVLM-Tiny ($1.3$B), being $5.5\times$ smaller, surpasses LLaVA-$7$B by $5.85\%$ on FFT.

Table \ref{tab:aokvqa} shows A-OKVQA results with a similar trend: SmoGVLM outperforms LLaVA across model sizes.
These findings highlight the benefits of injecting structured knowledge in VLMs.
However, ConceptNet's focus on scientific facts limits its effectiveness for A-OKVQA, which relies more on commonsense and social reasoning.

\begin{table}[t]
    \centering
    \small
    \begin{tabular}{l|c|c}
        \toprule
        \textbf{Sub-graph} & \textbf{Time} & \textbf{Similarity of top-$200$} \\
        \textbf{extraction method} & \textbf{Taken} & \textbf{triples with Answer} \\
        \midrule
        QA-GNN   & 7 hrs      & 0.232 $\pm$ 0.10 \\
        Proposed & \textbf{40 min} & \textbf{0.312 $\pm$ 0.14} \\
        \bottomrule
    \end{tabular}
    \caption{
        Comparison of sub-graph extraction methods.
        Our strategy is much faster and yields higher answer similarity.
    }
    \label{tab:subgraph}
\end{table}

\begin{table}[t]
    \centering
    \small
    \begin{tabular}{l|c|c|c}
        \toprule
        \textbf{Model} & \textbf{Size} & \textbf{Method} & \textbf{Acc. (\%)} \\
        \midrule
        LLaVA-Tiny     & $1.3$B        & LoRA            & 53.22              \\
        LLaVA-Tiny     & $1.3$B        & FFT             & 62.72              \\
        SmoGVLM-Tiny   & $1.3$B        & LoRA            & \textbf{69.46}              \\
        SmoGVLM-Tiny   & $1.3$B        & FFT             & \textbf{75.27}              \\
        \midrule
        LLaVA          & $7$B          & LoRA            & 69.41              \\
        LLaVA          & $7$B          & FFT             & 69.42              \\
        SmoGVLM        & $7$B          & LoRA            & \textbf{79.27}              \\
        SmoGVLM        & $7$B          & FFT             & \textbf{86.87}              \\
        \midrule
        LLaVA          & $13$B         & LoRA            & 81.58              \\
        LLaVA          & $13$B         & FFT             & 87.1               \\
        SmoGVLM        & $13$B         & LoRA            & 83.94              \\
        SmoGVLM        & $13$B         & FFT             & 88.02              \\
        \bottomrule
    \end{tabular}
    \caption{
        Performance on ScienceQA test split.
    }
    \label{tab:scienceqa}
\end{table}

\begin{table}[h]
    \centering
    \small
    \begin{tabular}{l|c|c|c}
        \toprule
        \textbf{Model} & \textbf{Size} & \textbf{Method} & \textbf{Acc. (\%)} \\
        \midrule
        LLaVA-Tiny     & $1.3$B        & FFT             & 70.4               \\
        SmoGVLM-Tiny   & $1.3$B        & FFT             & \textbf{70.7}      \\
        \midrule
        LLaVA          & $7$B          & FFT             & 79.3               \\
        SmoGVLM        & $7$B          & FFT             & \textbf{79.5}      \\
        \bottomrule
    \end{tabular}
    \caption{
        Performance on A-OKVQA validation split. 
    }
    \label{tab:aokvqa}
\end{table}

\section{Discussion}
\label{sec:discussion}

\subsection{Analysis of extracted triples}
We evaluate the relevance of retrieved triples by measuring their similarity with ground-truth answers\footnote{We use https://huggingface.co/sentence-transformers/all-mpnet-base-v2 to compute sentence embeddings.}.
Each triple is verbalized in \texttt{subject-relation-object} form and compared with the correct answer using cosine similarity. 
For each sample, we average these similarities to obtain a proximity score, and then take the mean over all samples to quantify overall relevance.
Table \ref{tab:subgraph} clearly demonstrates the efficiency and relevance advantages of our approach.

\begin{figure}
    \centering
    \includegraphics[width=\linewidth]{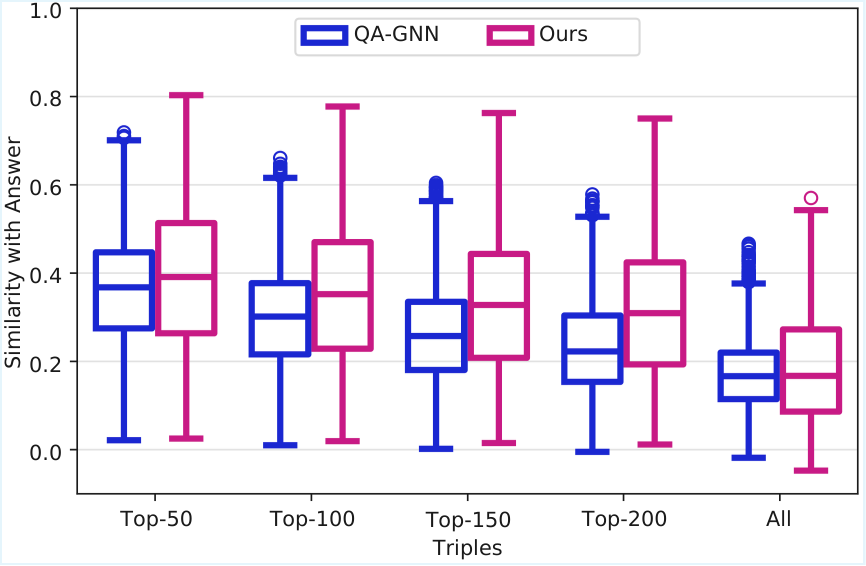}
    \caption{
        Similarity of top-$k$ triples to the correct answer.
    }
    \label{fig:triplesim}
\end{figure}

Figure \ref{fig:triplesim} further reports similarities as the number of retained triples increases.
We observe steadily higher relevance with larger sub-graphs, but the gain saturates beyond $200$ triples, suggesting that compact graphs capture most of the useful knowledge. 
Compared to QA-GNN, our method achieves higher similarity, and an order of magnitude faster, showing that efficiency and quality need not be traded off.

\subsection{Effect of sub-graph size}
Table \ref{tab:nodes} shows that performance improves steadily as the sub-graph grows, but saturates around $200$ nodes. 
This suggests that relatively small graphs already provide sufficient signal for reasoning, supporting the efficiency of our lightweight extraction strategy.
We hypothesize that larger graphs may only add noise with little benefit, and reserve a more comprehensive examination of this phenomenon for future research.

\begin{table}[h]
    \centering
    \small
    \begin{tabular}{l|c|c|c|c}
        \toprule
        \textbf{Model} & \textbf{Size}         & \textbf{Method}      & \textbf{\# Nodes}  & \textbf{Acc. (\%)} \\
        \midrule
        LLaVA          & \multirow{4}{*}{$7$B} & \multirow{4}{*}{FFT} & -                  & 69.42              \\ 
        \cmidrule{1-1}\cmidrule{4-5}
                       &                       &                      & 50                 & 83.94              \\
        SmoGVLM        &                       &                      & 100                & 86.65              \\
                       &                       &                      & 200                & 86.87              \\
        \bottomrule
    \end{tabular}
    \caption{
        Impact of sub-graph size on accuracy. Small graphs are sufficient, with performance saturating beyond 200 nodes.
    }
    \label{tab:nodes}
\end{table}

\subsection{Qualitative samples}
\begin{figure}[t]
    \centering
    \includegraphics[width=\linewidth]{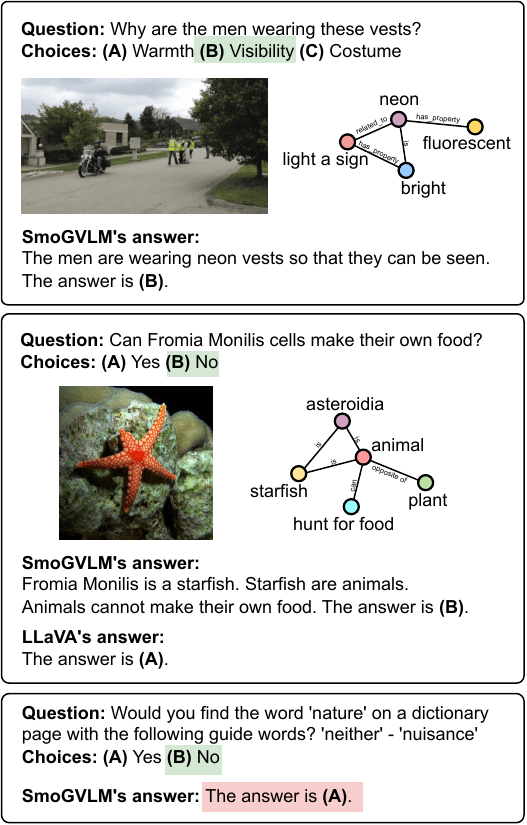}
    \caption{
        A few qualitative samples showing the strengths and limits of structured knowledge with VLMs.
    }
    \label{fig:samples}
\end{figure}

Figure \ref{fig:samples} presents some qualitative samples.
In the first case, we observe that the visual reasoning ability of VLMs remain intact, while the addition of the KG component enables the model to provide a grounded answer.
The second case shows how the availability of relevant triples allows SmoGVLM to predict the correct answer where LLaVA fails. 
Finally, the third example is a dictionary question, where structured KG-based augmentation cannot help. 
Together, these examples show how structured knowledge can enhance reasoning while also revealing tasks where it provides little benefit.

\section{Conclusion}
\label{sec:conclusion}
We introduce SmoGVLM, a small, graph-enhanced VLM for knowledge-intensive question answering.
By incorporating structured KGs with GNNs, SmoGVLM enables smaller models to outperform larger baselines.
This highlights a promising path towards efficient, knowledge-grounded intelligence.
Despite these gains, limitations remain.
KGs like ConceptNet offer incomplete coverage, and our fusion strategy is based on simple concatenation.
In the future, we plan to explore richer fusion mechanisms and broader benchmarks.

\bibliographystyle{IEEEbib}
\bibliography{refs}

\begin{thebibliography}{10}

\bibitem{liu2023visual}
Haotian Liu, Chunyuan Li, Qingyang Wu, and Yong~Jae Lee,
\newblock ``Visual instruction tuning,'' 2023.

\bibitem{li2023blip}
Junnan Li, Dongxu Li, Silvio Savarese, and Steven Hoi,
\newblock ``Blip-2: Bootstrapping language-image pre-training with frozen image encoders and large language models,''
\newblock {\em arXiv preprint arXiv:2301.12597}, 2023.

\bibitem{speer2017conceptnet}
Robyn Speer, Joshua Chin, and Catherine Havasi,
\newblock ``Conceptnet 5.5: An open multilingual graph of general knowledge,'' 2017.

\bibitem{yasunaga2021qa}
Michihiro Yasunaga, Hongyu Ren, Antoine Bosselut, Percy Liang, and Jure Leskovec,
\newblock ``{QA}-{GNN}: Reasoning with language models and knowledge graphs for question answering,''
\newblock in {\em Proceedings of the 2021 Conference of the North American Chapter of the Association for Computational Linguistics: Human Language Technologies}, Online, June 2021, pp. 535--546, Association for Computational Linguistics.

\bibitem{lu2022learn}
Pan Lu, Swaroop Mishra, Tony Xia, Liang Qiu, Kai-Wei Chang, Song-Chun Zhu, Oyvind Tafjord, Peter Clark, and Ashwin Kalyan,
\newblock ``Learn to explain: Multimodal reasoning via thought chains for science question answering,''
\newblock in {\em Advances in Neural Information Processing Systems}, Alice~H. Oh, Alekh Agarwal, Danielle Belgrave, and Kyunghyun Cho, Eds., 2022.

\bibitem{aokvqa}
Dustin Schwenk, Apoorv Khandelwal, Christopher Clark, Kenneth Marino, and Roozbeh Mottaghi,
\newblock ``A-okvqa: A benchmark for visual question answering using world knowledge,''
\newblock {\em arXiv}, 2022.

\bibitem{dai2023instructblip}
Wenliang Dai, Junnan Li, Dongxu Li, Anthony Tiong, Junqi Zhao, Weisheng Wang, Boyang Li, Pascale~N Fung, and Steven Hoi,
\newblock ``Instructblip: Towards general-purpose vision-language models with instruction tuning,''
\newblock {\em Advances in neural information processing systems}, vol. 36, pp. 49250--49267, 2023.

\bibitem{zhu2023minigpt}
Deyao Zhu, Jun Chen, Xiaoqian Shen, Xiang Li, and Mohamed Elhoseiny,
\newblock ``Minigpt-4: Enhancing vision-language understanding with advanced large language models,''
\newblock {\em arXiv preprint arXiv:2304.10592}, 2023.

\bibitem{liu2024survey}
Hanchao Liu, Wenyuan Xue, Yifei Chen, Dapeng Chen, Xiutian Zhao, Ke~Wang, Liping Hou, Rongjun Li, and Wei Peng,
\newblock ``A survey on hallucination in large vision-language models,''
\newblock {\em arXiv preprint arXiv:2402.00253}, 2024.

\bibitem{mondal2024kam}
Debjyoti Mondal, Suraj Modi, Subhadarshi Panda, Rituraj Singh, and Godawari~Sudhakar Rao,
\newblock ``Kam-cot: Knowledge augmented multimodal chain-of-thoughts reasoning,''
\newblock in {\em Proceedings of the AAAI Conference on Artificial Intelligence}, 2024, vol.~38, pp. 18798--18806.

\bibitem{wang2024knowledge}
Song Wang, Yaochen Zhu, Haochen Liu, Zaiyi Zheng, Chen Chen, and Jundong Li,
\newblock ``Knowledge editing for large language models: A survey,''
\newblock {\em ACM Computing Surveys}, vol. 57, no. 3, pp. 1--37, 2024.

\bibitem{schlichtkrull2018modeling}
Michael Schlichtkrull, Thomas~N Kipf, Peter Bloem, Rianne Van Den~Berg, Ivan Titov, and Max Welling,
\newblock ``Modeling relational data with graph convolutional networks,''
\newblock in {\em European semantic web conference}. Springer, 2018, pp. 593--607.

\bibitem{lin2019kagnet}
Bill~Yuchen Lin, Xinyue Chen, Jamin Chen, and Xiang Ren,
\newblock ``{K}ag{N}et: Knowledge-aware graph networks for commonsense reasoning,''
\newblock in {\em Proceedings of the 2019 Conference on Empirical Methods in Natural Language Processing and the 9th International Joint Conference on Natural Language Processing (EMNLP-IJCNLP)}, Hong Kong, China, Nov. 2019, pp. 2829--2839, Association for Computational Linguistics.

\bibitem{yasunaga2022deep}
Michihiro Yasunaga, Antoine Bosselut, Hongyu Ren, Xikun Zhang, Christopher~D Manning, Percy~S Liang, and Jure Leskovec,
\newblock ``Deep bidirectional language-knowledge graph pretraining,''
\newblock {\em Advances in Neural Information Processing Systems}, vol. 35, pp. 37309--37323, 2022.

\bibitem{touvron2023llama}
Hugo Touvron, Louis Martin, Kevin Stone, Peter Albert, Amjad Almahairi, Yasmine Babaei, Nikolay Bashlykov, Soumya Batra, Prajjwal Bhargava, Shruti Bhosale, et~al.,
\newblock ``Llama 2: Open foundation and fine-tuned chat models,''
\newblock {\em arXiv preprint arXiv:2307.09288}, 2023.

\bibitem{DBLP:conf/icml/RadfordKHRGASAM21}
Alec Radford, Jong~Wook Kim, Chris Hallacy, Aditya Ramesh, Gabriel Goh, Sandhini Agarwal, Girish Sastry, Amanda Askell, Pamela Mishkin, Jack Clark, Gretchen Krueger, and Ilya Sutskever,
\newblock ``Learning transferable visual models from natural language supervision,''
\newblock in {\em Proceedings of the 38th International Conference on Machine Learning, {ICML} 2021, 18-24 July 2021, Virtual Event}, Marina Meila and Tong Zhang, Eds. 2021, vol. 139 of {\em Proceedings of Machine Learning Research}, pp. 8748--8763, {PMLR}.

\bibitem{busbridge2019relational}
Dan Busbridge, Dane Sherburn, Pietro Cavallo, and Nils~Y. Hammerla,
\newblock ``Relational graph attention networks,'' 2019.

\bibitem{DBLP:conf/iclr/KipfW17}
Thomas~N. Kipf and Max Welling,
\newblock ``Semi-supervised classification with graph convolutional networks,''
\newblock in {\em 5th International Conference on Learning Representations, {ICLR} 2017, Toulon, France, April 24-26, 2017, Conference Track Proceedings}. 2017, OpenReview.net.

\end{thebibliography}

\end{document}